\documentclass{article}

\usepackage[preprint]{neurips_2025}
\usepackage{graphicx}
\usepackage{booktabs}
\usepackage{siunitx}
\usepackage{adjustbox}
\usepackage{float}
\usepackage{subcaption}
\usepackage{wrapfig}
\usepackage{multirow}

\usepackage[utf8]{inputenc} 
\usepackage[T1]{fontenc}    
\usepackage{hyperref}       
\usepackage{url}            
\usepackage{booktabs}       
\usepackage{amsfonts}       
\usepackage{nicefrac}       
\usepackage{microtype}      
\usepackage{xcolor,soul}         
\usepackage{amsmath}

\title{Rethinking Perplexity: Revealing the Impact of Input Length on Perplexity Evaluation in LLMs}

\author{
  Letian Cheng$^{1}$,
  Junyan Wang$^{2}$,
  Yan Gao$^{3}$,
  Elliott Wen$^{4}$,
  Ting Dang$^{1}$,
  Hong Jia$^{4,1}$ \\
  $^{1}$University of Melbourne, 
  $^{2}$University of Adelaide, \\
  $^{3}$University of Cambridge,
  $^{4}$University of Auckland \\
  \texttt{letiancheng@student.unimelb.edu.au},
  \texttt{junyan.wang@adelaide.edu.au},\\
  \texttt{yg381@cam.ac.uk},
  \texttt{elliott.wen@auckland.ac.nz},\\
  \texttt{ting.dang@unimelb.edu.au},
  \texttt{hong.jia@auckland.ac.nz}
}

\begin{document}

\maketitle

\begin{abstract}
Perplexity is a widely adopted metric for assessing the predictive quality of large language models (LLMs) and often serves as a reference metric for downstream evaluations. However, recent evidence shows that perplexity can be unreliable, especially when irrelevant long inputs are used, raising concerns for both benchmarking and system deployment. While prior efforts have employed selective input filtering and curated datasets, the impact of input length on perplexity has not been systematically studied from a systems perspective and input length has rarely been treated as a first-class system variable affecting both fairness and efficiency.
In this work, we close this gap by introducing LengthBenchmark, a system-conscious evaluation framework that explicitly integrates input length, evaluation protocol design, and system-level costs, evaluating representative LLMs under two scoring protocols (direct accumulation and fixed window sliding) across varying context lengths. Unlike prior work that focuses solely on accuracy-oriented metrics, LengthBenchmark additionally measures latency, memory footprint, and evaluation cost, thereby linking predictive metrics to deployment realities. We further incorporate quantized variants not as a main contribution, but as robustness checks, showing that length-induced biases persist across both full-precision and compressed models. This design disentangles the effects of evaluation logic, quantization, and input length, and demonstrates that length bias is a general phenomenon that undermines fair cross-model comparison.
Our analysis yields two key observations: (i) sliding window evaluation consistently inflates performance on short inputs, and (ii) both full-precision and quantized models appear to realise gains as the evaluated segment length grows. Taken together, these findings establish input length as a fundamental dimension in LLM benchmarking, highlighting the limitations of perplexity as currently formulated and motivating the need for length-aware, system-conscious evaluation protocols that better guide real-world deployment.
\end{abstract}

\section{Introduction}


The remarkable advancements in large language models (LLMs) have led to their widespread adoption in applications ranging from natural language understanding to content generation. As these models grow in scale and complexity, the need for reliable and cost-effective evaluation metrics has become increasingly critical. Perplexity, derived from cross-entropy loss, has long been a cornerstone for assessing language models due to its probabilistic interpretation and ease of computation~\cite{jelinek1977perplexity, bengio2003neural}.

However, perplexity in its current form suffers from fundamental limitations: it does not always correlate with downstream task performance~\cite{luden2024beyond, DBLP:conf/emnlp/XuG0BS24}, and its reliability deteriorates as input length increases~\cite{DBLP:conf/iclr/0002HTZF24,kuribayashi-etal-2021-lower, DBLP:conf/acl/LevyJG24}. Prior efforts such as LongPPL~\cite{DBLP:conf/iclr/FangWLZJGD025} and LongBench~\cite{DBLP:conf/acl/BaiLZL0HDLZHDTL24, DBLP:conf/acl/BaiTZ0WLCX0D0L25} attempted to mitigate this by curating datasets or filtering inputs, yet they largely treated input length as a nuisance factor, rather than a first-class variable that directly shapes both predictive metrics and system-level efficiency.

In this work, we close this gap by introducing LengthBenchmark, a system-conscious evaluation framework that explicitly integrates input length, evaluation protocols, and deployment-relevant costs. LengthBenchmark evaluates representative LLMs under two scoring protocols (direct accumulation, fixed window sliding) across varying context lengths, and incorporates quantized variants to reflect efficiency-driven deployments. Unlike prior studies, our framework measures not only predictive quality (e.g., perplexity, accuracy) but also system-level metrics such as memory footprint, inference latency, and evaluation cost, thereby linking evaluation reliability to deployment realities.

Our analysis yields two key insights: (i) protocol design (sliding vs. non-sliding) introduces systematic bias, particularly at long input lengths, and (ii) length-induced trends persist across both full-precision and quantized models, underscoring the robustness of this phenomenon. Taken together, these findings establish input length as a fundamental dimension of LLM evaluation, and motivate the need for length-aware, system-conscious benchmarking practices that better guide real-world deployment.

Our contributions are three-fold:
\begin{itemize}
\item Length as a system variable. We systematically demonstrate that input length is not a neutral parameter but a fundamental factor that shapes perplexity outcomes and benchmarking fairness.
\item Protocol-level bias. We reveal that common evaluation protocols (sliding vs. non-sliding) introduce systematic and length-dependent biases, which distort model comparison.
\item System-conscious evaluation. We propose LengthBenchmark, the first framework that unifies input length variation, evaluation protocols, and system-level costs (latency, memory, evaluation overhead). While we also include quantized variants, their role is to confirm that length-induced biases persist under real-world compression settings, underscoring the robustness of our findings.
\end{itemize}
\vspace{-0.5em}


\section{Length-Induced Biases in Perplexity Evaluation}\label{sec:2}
Current perplexity evaluation methods commonly employ a sliding window to assess language model performance. This method segments input text into overlapping windows of fixed length $w$, calculating the model's loss separately for each window. Given a sequence of tokens $x_1, x_2, \ldots, x_T$, the sliding window method creates overlapping subsequences $x_i, x_{i+1}, \ldots, x_{i+w-1}$ for $i = 1, 2, \ldots, T-w+1$. The loss for each window is computed as $\mathcal{L}_i = -\frac{1}{w} \sum_{j=i}^{i+w-1} \log p(x_j | x_{<j})$. The total loss is then aggregated across all windows as $\mathcal{L} = \frac{1}{T-w+1} \sum_{i=1}^{T-w+1} \mathcal{L}_i$, and the average loss is converted into a perplexity score using $\text{PPL} = \exp(\mathcal{L})$.
\begin{wrapfigure}{r}{0.35\textwidth}
  \vspace{-1em}
  \centering
  \includegraphics[width=0.35\textwidth]{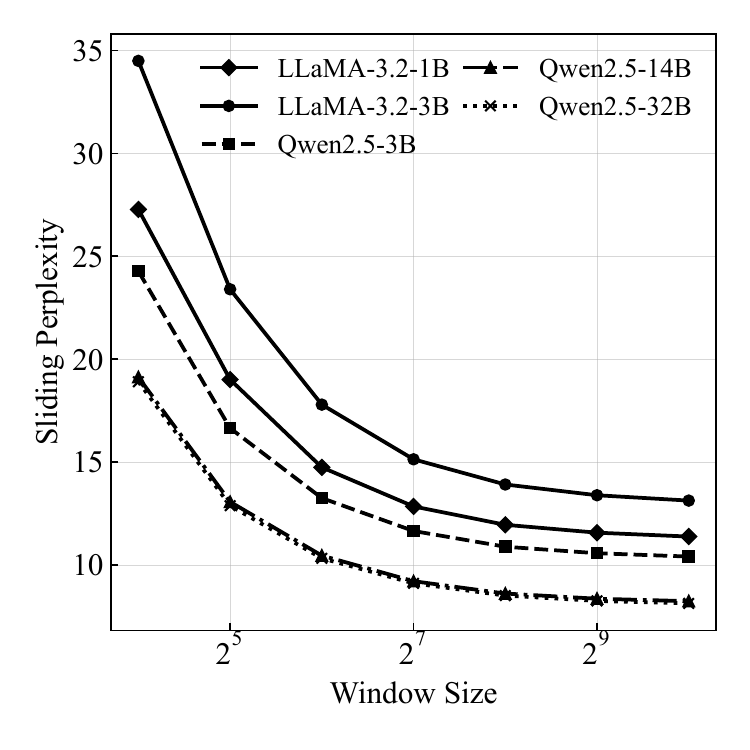}
  \vspace{-1em}
  \caption{Effect of Sliding Window Size on Perplexity (C4, SeqLen=2048)}
  \label{fig:1}
  \vspace{-3.3em}
\end{wrapfigure}
This design enables evaluation under limited context windows while covering long sequences. However, it introduces critical limitations: results are highly sensitive to the chosen window size $w$, and the method often underestimates model capability on long inputs. We detail these issues below.



\subsection{Sensitivity Within Sliding Window Evaluation}\label{subsec:1}

Most existing perplexity evaluations, whether handcrafted (e.g., SmoothQuant~\cite{DBLP:conf/icml/XiaoLSWDH23}) or toolkit-based (e.g., lm-eval-harness~\cite{eval-harness}), compute loss over overlapping sliding windows. However, the choice of window size has a significant effect on the reported score. As shown in Figure~\ref{fig:1}, increasing the window from 16 to 1024 tokens reduces perplexity by 57–62\% across LLaMA-3.2 and Qwen models of different scales. This demonstrates that sliding-window evaluation systematically undervalues long-context modelling ability, with comparable distortion across model sizes.

This behavior resembles a scaling-law effect: larger windows expose the model to more contextual information, enabling it to exploit long-range dependencies more effectively, especially in higher-capacity models. This sensitivity means that reported perplexity depends as much on evaluation protocol design as on model quality, raising concerns for fair benchmarking and system-level comparisons.
\subsection{Protocol Bias (Sliding vs. Non-Sliding)}\label{subsec:2}
Since perplexity is defined as the exponential of the average negative log-likelihood, the normalization denominator strongly affects the reported score. To examine this effect, we compare two strategies:
(i). Sliding window: loss is computed within fixed windows of 1k (i.e., 1,024) tokens and averaged.
(ii). Non-sliding (direct accumulation): loss is accumulated across the full sequence without segmentation.

To examine the impact of different size of input tokens, we evaluate LLaMA-3.2 and Qwen-2.5 models on the C4 dataset~\cite{DBLP:conf/emnlp/DodgeSMAIGM021} with input lengths of 1,024–8,192 tokens. Results in Table~\ref{tab:non-sliding-vs-sliding} show that non-sliding consistently achieves higher accuracy and often lower perplexity than sliding, with differences widening as sequence length and model size increase. This indicates that sliding evaluation systematically underestimates long-context performance, particularly for larger models.
\newcommand{\better}[1]{\textbf{#1} $\uparrow$} 
\newcommand{\worse}[1]{\textbf{#1} $\downarrow$} 
\begingroup
\begin{table}
  \centering
  \caption{Non-Sliding vs.\ Sliding evaluation across sequence lengths. 
  $\Delta$PPL $=$ Sliding$-$Non\mbox{-}Sliding (lower is better), 
  $\Delta$Acc $=$ Non\mbox{-}Sliding$-$Sliding (higher is better). 
  Arrows indicate which method is better (↑ Non-Sliding, ↓ Sliding).}
  \label{tab:non-sliding-vs-sliding}
  \vspace{\baselineskip}
  \begin{adjustbox}{max width=\textwidth}
  \begin{tabular}{
    l
    S[table-format=4.0]  
    S[table-format=2.4]  
    S[table-format=2.4]  
    S[table-format=+1.4] 
    S[table-format=2.4]  
    S[table-format=2.4]  
    S[table-format=+1.4] 
  }
  \toprule
  \textbf{Model} & \textbf{Seq Len} & \textbf{PPL (Non-Sliding)} & \textbf{PPL (Sliding)} & {$\boldsymbol{\Delta}$\textbf{PPL}} & \textbf{Acc\% (Non-Sliding)} & \textbf{Acc\% (Sliding)} & {$\boldsymbol{\Delta}$\textbf{Acc}} \\
  \midrule
  \multirow{4}{*}{LLaMA-3.2-3B} & 1024 & 15.7299 & 15.7140 & \worse{-0.0159} & 43.6972 & 43.0966 & \better{+0.6006} \\
               & 2048 & 14.6484 & 15.7140 & \better{+1.0657} & 44.9319 & 43.0966 & \better{+1.8353} \\
               & 4096 & 13.8876 & 15.7140 & \better{+1.8264} & 45.7912 & 43.0966 & \better{+2.6946} \\
               & 8192 & 12.9255 & 15.7140 & \better{+2.7885} & 47.1186 & 43.0966 & \better{+4.0220} \\
  \midrule
  \multirow{4}{*}{LLaMA-3.2-1B} & 1024 & 13.8083 & 13.9705 & \better{+0.1622} & 45.5431 & 44.5556 & \better{+0.9875} \\
               & 2048 & 12.9397 & 13.9705 & \better{+1.0307} & 46.7063 & 44.5556 & \better{+2.1506} \\
               & 4096 & 12.0809 & 13.9705 & \better{+1.8896} & 47.9805 & 44.5556 & \better{+3.4249} \\
               & 8192 & 11.2433 & 13.9705 & \better{+2.7271} & 49.3179 & 44.5556 & \better{+4.7623} \\
  \midrule
  \multirow{4}{*}{Qwen2.5-3B} & 1024 & 13.0729 & 13.0541 & \worse{-0.0187} & 46.2463 & 45.6544 & \better{+0.5919} \\
               & 2048 & 12.2327 & 13.0541 & \better{+0.8214} & 47.4003 & 45.6544 & \better{+1.7459} \\
               & 4096 & 11.2248 & 13.0541 & \better{+1.8293} & 49.0313 & 45.6544 & \better{+3.3769} \\
               & 8192 & 10.3074 & 13.0541 & \better{+2.7467} & 50.5993 & 45.6544 & \better{+4.9449} \\
  \midrule
  \multirow{4}{*}{Qwen2.5-14B} & 1024 & 10.1741 & 10.1212 & \worse{-0.0529} & 49.3371 & 48.6627 & \better{+0.6744} \\
               & 2048 &  9.5452 & 10.1212 & \better{+0.5760} & 50.5184 & 48.6627 & \better{+1.8557} \\
               & 4096 &  8.8256 & 10.1212 & \better{+1.2956} & 52.0906 & 48.6627 & \better{+3.4279} \\
               & 8192 &  8.1612 & 10.1212 & \better{+1.9600} & 53.6575 & 48.6627 & \better{+4.9948} \\
  \midrule
  \multirow{4}{*}{Qwen2.5-32B} & 1024 & 10.0648 & 10.0853 & \better{+0.0205} & 49.6297 & 48.9665 & \better{+0.6632} \\
               & 2048 &  9.4372 & 10.0853 & \better{+0.6481} & 50.8349 & 48.9665 & \better{+1.8684} \\
               & 4096 &  8.7135 & 10.0853 & \better{+1.3718} & 52.4354 & 48.9665 & \better{+3.4689} \\
               & 8192 &  8.0536 & 10.0853 & \better{+2.0317} & 54.0076 & 48.9665 & \better{+5.0411} \\
  \bottomrule
  \end{tabular}
  \end{adjustbox}
\end{table}
\endgroup

Overall, Sections \ref{subsec:1} and \ref{subsec:2} reveal two related issues: (i) instability within sliding evaluation due to window size sensitivity, and (ii) systematic bias between sliding and non-sliding protocols, which magnifies with input length and model scale. These findings highlight that input length is not a neutral parameter but a fundamental system constraint: it could distort predictive metrics and undermines fair cross-model benchmarking. Motivated by this, we introduce LengthBenchmark, a framework that integrates input length with system-level concerns such as quantization, inference latency, and memory footprint, to enable more realistic and efficiency-aware evaluation of LLMs.
\section{LengthBenchmark: A Framework for Length-Aware Evaluation}
The analyses in Section \ref{sec:2} revealed that input length interacts with perplexity in ways that introduce instability and bias. However, these findings were limited to perplexity alone, leaving open the question of how length effects manifest under more realistic deployment conditions. To address this, we introduce LengthBenchmark, a framework that systematically integrates input length, evaluation protocols, and quantization methods to provide a more comprehensive and system-relevant assessment of LLM performance

\textbf{Sequence Lengths.} Instead of relying on a single, fixed sequence length, we evaluate models across a range of input lengths (i.e., 1K-8K tokens). This design not only reveals length-induced biases but also reflects real-world usage, where deployment scenarios involve diverse sequence lengths.

\textbf{Evaluation Protocols.} We compare two primary perplexity calculation methods: the standard sliding window approach and a non-sliding (direct accumulation) method. This dual-protocol evaluation enables a direct comparison between the conventional method and a more straightforward approach that avoids the biases introduced by windowing.

\textbf{Quantization Methods.} Since real-world deployment of LLMs usually relies on quantization for efficiency, LengthBenchmark integrates widely adopted methods including AWQ~\cite{DBLP:conf/mlsys/0002TTYCWXDG024}, SmoothQuant~\cite{DBLP:conf/icml/XiaoLSWDH23}, GPTQ~\cite{DBLP:journals/corr/abs-2210-17323}, and HQQ~\cite{mobiusml2023hqq}. By measuring their performance under varying input lengths, the benchmark exposes how quantization interacts with sequence length, an aspect overlooked in prior evaluations. This ensures that both full-precision and quantized models are compared fairly under realistic system constraints.

\section{Results and Findings}

\subsection{Length-induced Bias in Full Precision}
As discussed in Section~\ref{sec:2}, sliding-window evaluation is highly sensitive to window size and systematically underestimates long-context performance compared to non-sliding protocols. Using LengthBenchmark, we show these effects across multiple models (ref. Figure~\ref{fig:1}, Table~\ref{tab:non-sliding-vs-sliding}), establishing a consistent basis for the quantization analysis that follows.
\subsection{Robustness under quantizaion}
\begin{figure}[t]
    \centering
    \includegraphics[width=0.98\textwidth]{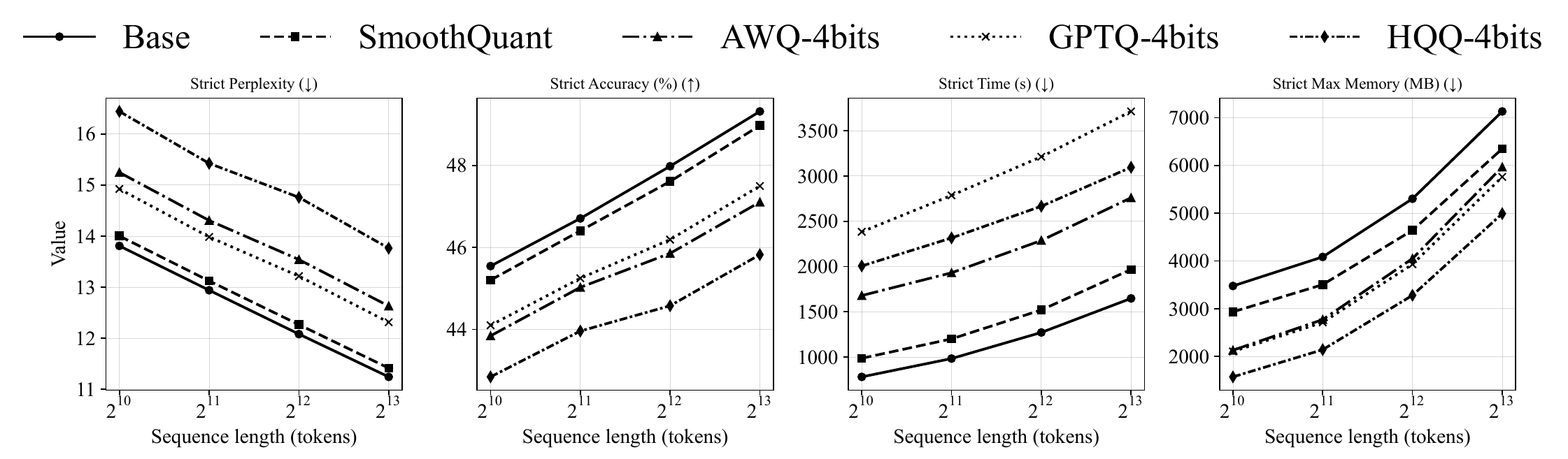}
    \caption{Comparison of quantization methods across different sequence lengths under strict evaluation.}
    \label{fig:3}
\end{figure}

Real-world deployment of LLMs usually relies on quantization to meet efficiency constraints. As such, it is important to understand whether length-induced biases persist under quantized models. Using LengthBenchmark, we evaluated AWQ, SmoothQuant, GPTQ, and HQQ on LLaMA-3.2-1B across multiple sequence lengths (Figure~\ref{fig:3}).

Our results show that all quantization methods benefit from longer sequences (achieving lower perplexity), but the degree of improvement varies: SmoothQuant remains relatively stable, while AWQ and HQQ exhibit more pronounced gains. For example, AWQ-4bit perplexity drops from 15.2 at 1k tokens to 12.6 at 8k (a 17\% reduction), while SmoothQuant decreases more moderately from 14.0 to 11.4. HQQ-4bit shows the steepest relative gain, from 16.4 to 13.8 (16\% reduction), albeit with consistently lower accuracy compared to other methods (42.8\% at 1k to 45.8\% at 8k). In contrast, SmoothQuant maintains relatively stable accuracy (45.2\% to 49.0\%) and the lowest memory footprint at longer lengths (6.3GB vs. 7.1GB for the baseline). 

Latency trends reveal a trade-off: AWQ is faster than GPTQ at shorter inputs (1.7s vs. 2.4s at 1k) but the gap narrows at 8k (2.8s vs. 3.7s). HQQ offers the smallest memory usage (1.6GB at 1k, 5.0GB at 8k) but incurs higher latency than SmoothQuant. These variations imply that quantization interacts with input length, complicating fair efficiency–accuracy trade-offs and underscoring the importance of benchmarking under diverse length settings.


\subsection{System-Level Implications}
The results across both full-precision and quantized models demonstrate that input length shapes both perplexity, accuracy, memory footprint and latency. LengthBenchmark highlights the need for length-aware and system-conscious evaluation protocols, ensuring that benchmarking practices fairly capture both predictive quality and system efficiency across realistic usage scenarios.

\section{Conclusion and Future Work}
Perplexity remains a widely used metric for evaluating large language models, yet our study shows that its reliability is fundamentally constrained by input length. Through LengthBenchmark, we systematically demonstrated that length effects manifest both within evaluation protocols (sliding vs. non-sliding) and across quantization strategies. These observations indicate that input length is not a minor methodological detail but a core system variable that influences predictive quality, efficiency claims, and benchmarking fairness. Our results underscore the need for length-aware, system-conscious evaluation practices. We envision more standardized benchmarks such as LengthBenchmark will ensure not only fairer reporting of model quality, but also more reliable guidance for deployment decisions, scalability assessments, and efficiency trade-offs. As future work, we plan to extend our framework to cover more diverse datasets and model families, and to refine evaluation protocols that further reduce length-induced bias. 


\newpage
{
\small
\bibliographystyle{unsrt}  
\bibliography{references}  
}

\end{document}